%% file: root.tex
\documentclass[letterpaper, 10 pt, conference]{ieeeconf}

\IEEEoverridecommandlockouts

\usepackage[T1]{fontenc}
\usepackage[utf8]{inputenc}

\usepackage{etextools}
\usepackage{amssymb}
\usepackage{bm}
\usepackage{float}
\usepackage{makeidx}
\usepackage{amsmath}
\usepackage{algorithm}
\usepackage{bbm}
\usepackage{epsfig}
\usepackage{epsf}
\usepackage{psfrag}
\usepackage{verbatim}
\let\labelindent\relax
\usepackage{enumitem}
\usepackage{color}
\usepackage{multirow}
\usepackage{tabularx}
\usepackage{booktabs}
\usepackage{svg}
\usepackage{adjustbox}
\usepackage[tight,footnotesize]{subfigure}
\usepackage{array}
\usepackage{soul}
\usepackage{footnote}
\usepackage{cite}
\usepackage{dblfloatfix}

\usepackage{color, colortbl}
\usepackage{graphicx}
\graphicspath{{Figures}}
\DeclareGraphicsExtensions{.pdf,.png}
\usepackage{bigstrut}
\usepackage[english]{babel}
\usepackage{csquotes}
\usepackage[T1]{fontenc}
\usepackage{algorithm, setspace}
\usepackage{algpseudocode}
\usepackage{url}
\usepackage{multirow}
\usepackage{float}
\usepackage{xcolor}
\usepackage{hyperref}
 \hypersetup{
     colorlinks=true,
     linkcolor=orange,
     filecolor=orange,
     citecolor=orange,      
     urlcolor=orange,
     }

\newcommand{\p}[1]{\smallskip \noindent \textbf{{#1}.}}
\newcommand{\eq}[1]{Equation~(\ref{eq:#1})}
\newcommand{\fig}[1]{Figure~\ref{fig:#1}}

\usepackage{balance}

\title{\LARGE
Many-RRT$^\star$: Robust Joint-Space Trajectory Planning for Serial Manipulators
}

\author{
    Theodore M. Belmont$^1$, Benjamin A. Christie$^{1,2}$, and Anton Netchaev$^1$
    \thanks{
    This study was conducted for the US Army Engineer Research and Development Center (ERDC), as part of the Civil Works Research and Development area within the Next Generation Water Resources Infrastructure, under Funding Account Code U453885; AMSCO Code 008329. The views, opinions, findings, and conclusions reflected in this publication are solely those of the authors and do not necessarily represent the official policy or position of the U.S. Army Engineer Research and Development Center or the U.S. Army Corps of Engineers. 
    }
    \thanks{
    $^1$are with the USACE ERDC, Information Technology Lab (\href{https://www.erdc.usace.army.mil/Locations/ITL/}{ERDC}), Vicksburg, MS 39180, USA. \texttt{\{theodore.m.belmont, benjamin.a.christie, anton.netchaev\}@erdc.dren.mil}
    } 
    \thanks{
    $^2$is with the Collaborative Robotics Lab (\href{https://collab.me.vt.edu/}{Collab}), Dept. of Mechanical Engineering, Virginia Tech, Blacksburg, VA 24061.
    \texttt{benc00@vt.edu}
    }
}

\begin{document}
\maketitle

%%%%%%%%%%%%%%%%%%%%%%%%%%%%%%%%%%%%%%%%%%%%%%%%%%%%%%%%%%%%%%%%%%%%%%%%%%%%%%%%
\begin{abstract}
The rapid advancement of high degree-of-freedom (DoF) serial manipulators necessitates the use of swift, sampling-based motion planners for high-dimensional spaces. 
While sampling-based planners like the Rapidly-Exploring Random Tree (RRT) are widely used, planning in the manipulator's joint space presents significant challenges due to non-invertible forward kinematics.
A single task-space end-effector pose can correspond to multiple configuration-space states, creating a multi-arm bandit problem for the planner. 
In complex environments, simply choosing the wrong joint space goal can result in suboptimal trajectories or even failure to find a viable plan.
To address this planning problem, we propose Many-RRT$^\star$: an extension of RRT$^\star$-Connect that plans to multiple goals in parallel. 
By generating multiple IK solutions and growing independent trees from these goal configurations simultaneously alongside a single start tree, Many-RRT$^\star$ ensures that computational effort is not wasted on suboptimal IK solutions. 
This approach maintains robust convergence and asymptotic optimality. 
Experimental evaluations across robot morphologies and diverse obstacle environments demonstrate that Many-RRT$^\star$ provides higher quality trajectories ({44.5\% lower cost in the same runtime}) with a significantly higher success rate ({100\% vs. the next best of 1.6\%}) than previous RRT iterations without compromising on runtime performance.
\end{abstract}

\input{0_intro}
\input{1_related}
\input{2_problem}
\input{3_method}
\input{4_study}
\input{5_conclusion}

\balance
\bibliographystyle{IEEEtran}
\bibliography{IEEEabrv,bibtex}

\end{document}

%% file: 0_intro.tex
\section{Introduction}\label{sec:intro}

The rapid advancement of robotic platforms, such as high degree-of-freedom (DoF) manipulators or unmanned ground vehicles, has necessitated the use of swift, probabilistic motion planners for high-dimensional spaces. Sampling-Based Planners (SBPs), such as Probabilistic Roadmaps (PRMs, \cite{kavraki1996prm}) or Rapidly-Exploring Random Trees (RRTs, \cite{lavalle1998rapidly}) are the backbone of many public and commercial motion planning libraries. These approaches offer probabilistic completeness, and in some cases, asymptotic stability. They are highly scalable and simple to implement, making them the standard approach for waypoint-based planning. 

\begin{figure}[t]
    \centering
    \includegraphics[width=1\linewidth]{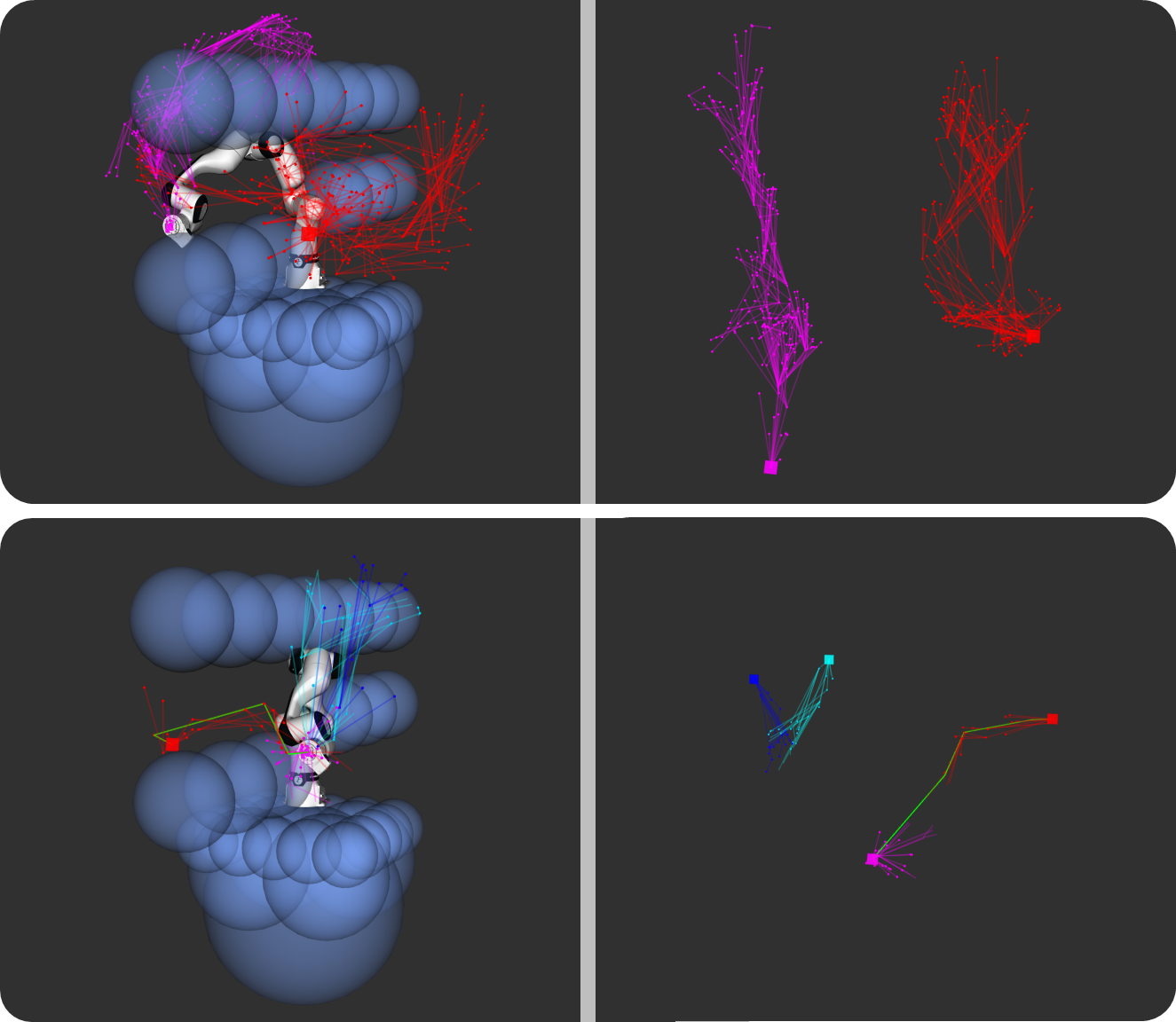}
    \caption{
    The trees created by RRT$^\star$-Connect (top) vs. Many-RRT$^\star$ (bottom) in the task-space (left) and joint-space (right). Despite having denser, more extensive trees, RRT$^\star$-Connect chooses a suboptimal \textit{goal} joint state from the preimage of the robot's forward kinematics. Since our method samples multiple goal configurations, we can solve for many possible plans in parallel, improving success rate and decreasing path costs.
    }
    \label{fig:front}
\end{figure}

However, forward kinematics for serial manipulators is non-invertable; a single end-effector pose can correspond to many unique joint configurations.
This property makes effective manipulator motion planning in the configuration-space non-trivial.
When planning in configuration-space, the manipulator encounters a multi-armed bandit problem where choosing the wrong goal joint configuration can lead to suboptimal motions.
In complex environments, this can result in planning failures --- even when viable plans exists --- simply because the planner chose the wrong goal state.

A naive solution to this problem is to solve for \textit{all} possible joint configurations resulting in the appropriate goal pose, but this is computationally infeasible. Even with manipulators with $6$-DoF, the preimage of specific goal poses can have uncountably infinite solutions. The problem exacerbates as the DoFs grow. 

We propose a solution to this problem with Many-RRT$^\star$, which is similar to RRT$^\star$-Connect \cite{klemm2015connect}, but extends multiple configuration-space goals in parallel.
This approach maintains the robust convergence and asymptotic-optimality guarantees of RRT$^\star$-Connect while offering significant advantages. 
By rephrasing the sampling-based algorithm as a search over the configuration space \textit{and} the preimage of the manipulator's forward kinematics, we are able to achieve a lower bound on path length in configuration space. 
As the number of samples from the forward kinematics preimage increases, Many-RRT$^\star$ changes from a \textit{locally} asymptotically-optimal motion planning algorithm from a \textit{globally} optimal one.

Overall, we make the following contributions:

\p{Formalizing Globally-Optimal Motion-Planners}
We define the context in which sampling-based motion planners (such as RRT$^\star$) exhibit \textit{global} asymptotic-optimality. When planning to known global optima (such as planning in task-space alone), existing planners exhibit global optimality. However, when planning in configuration-space, existing planners are \textit{not} globally optimal, due to the many-to-one behavior of the manipulator's forward kinematics.

\p{Solving One-to-Many Motion Plans}
We present a method for generating motion plans that samples from the preimage of the robot's forward kinematics and optimizes over the path length of each plan independently. 
With a single sample, our method is identical to RRT$^\star$-Connect. However, as the number of samples grows --- according to our formalization above --- the plan approaches \textit{global} optimality. Each plan is independent, so this approach is trivially parallelizable on modern hardware with virtually no difference in planning time when compared to start-of-the-art implementations. 

\p{Experimental Validation}
We test Many-RRT$^\star$ against the state-of-the-art single-query asymptotically-optimal motion planners RRT$^\star$ \cite{karaman2011rrtstar} and RRT$^\star$-Connect \cite{klemm2015connect} across robot morphologies and environments. We find that in the tested situations, Many-RRT$^\star$ is able to find motion plans with {44.5\% lower cost in the same runtime}. 
Further, we find that Many-RRT$^\star$ matches or exceeds the success rate of the baselines across environments and manipulators. 

%% file: 1_related.tex
\section{Related Works}\label{sec:related}

Motion planning in robotic manipulators remains an open challenge, despite its long history. Approaches range from swift sampling-based position planners \cite{klemm2015connect, gammell2014informed, noreen2016rrtreview, karaman2011rrtstar} to low-level receding-horizon control \cite{incremona2017mpc, rybus2017control}.
Complementing the position-sampling approaches, some have leveraged machine-learning to improve sampling quality in higher-dimensional spaces \cite{carvalho2023motion, huang2024diffusionseeder} or to condition samples on environment observations directly \cite{tianyi2022gan}.

\p{Path Planning}
Path planning has been a key interest in robotics since at least the mid-1970s. Early implementations focused on reducing the state representation to a 2D-plane and applying variants of Dijkstra's algorithm \cite{warren1993fast}, known to be (generally, see \cite{duan2025breaking}) theoretically optimal. Initial research established core techniques such as cell decomposition \cite{choset1998coverage, latombe1991exact}, visibility graphs \cite{janet1995essential, jiang1993finding}, and potential fields \cite{jiang1993finding, hwang1992potential}. As the computational power of consumer processors grew, sampling-based planners \cite{elbanhawi2014sampling, lavalle2000birrt} gained popularity. Sampling-based planners are now widely taken to be the de facto standard for high-level motion generators, specifically with the advent of bi-directional growth through RRT-Connect \cite{kuffner2000rrt} and asymptotic optimality in RRT$^\star$ \cite{karaman2011rrtstar}. 

\p{Extending RRT}
A plethora of RRT variants have been proposed to improve the performance of RRT-style algorithms. We are focused on two variants (and their combination) specifically. 
The first is RRT-Connect \cite{kuffner2000rrt}. This method extends the original RRT algorithm to create two trees, with one rooted at the starting state and the other rooted at the goal state. Convergence is achieved when the trees can connect. This modification improves the success rate of the original RRT algorithm, especially in the presence of obstacles.

The second is RRT$^\star$ \cite{karaman2011rrtstar}. This algorithm offers asymptotic optimality by modifying the parentage of nodes near each new sample. Convergence is achieved when a node of the tree is within a goal zone.
Klemm et. al \cite{klemm2015connect} combined these approaches to create RRT$^\star$-Connect, which leverages the robust convergence of RRT-Connect and the asymptotic optimality of RRT$^\star$. RRT$^\star$-Connect grows bi-directionally and iteratively improves both trees using the same rewiring logic as RRT$^\star$. However, although RRT$^\star$-Connect has asymptotic optimality for a particular start-goal tuple, it is subject to the same failures as other methods that do not reason over multiple goal configurations.

Following \cite{klemm2015connect}, there has been extensive research into using multiple RRTs in parallel. 
As discussed in \cite{huang2025prrtc}, there are three levels of parallelism. The nearest to our approach is those that create multiple trees in parallel \cite{huang2025prrtc, otte2013c, hidalgo2018quad}. However, these approaches do not plan to multiple \textit{goals} in parallel.
Using RRT to plan to multiple goals has been explored \cite{pereira2016path}, however, this work demonstrates planning to \textit{sequential} non-compulsory goal positions. 

% As we will show, selecting the optimal goal position in configuration-space is a multi-arm bandit problem, where selecting a suboptimal goal position can cause pessimization or even failures.

%% file: 2_problem.tex
\section{Problem Statement}\label{sec:problem}

\begin{figure}[t]
    \centering
    \includegraphics[width=1\linewidth]{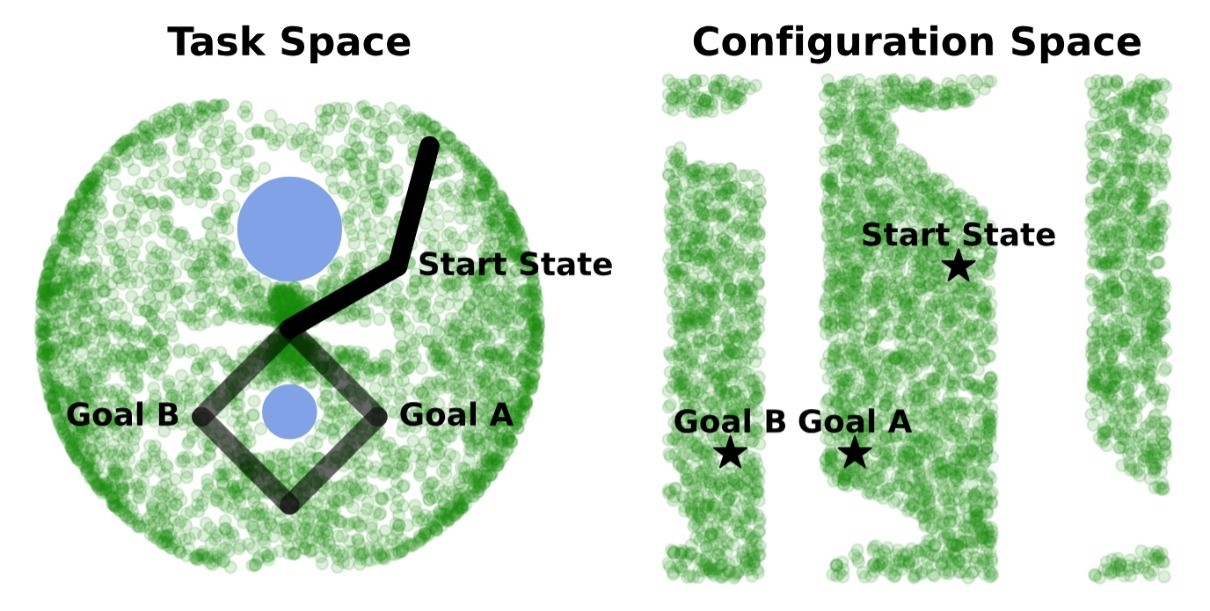}
    \caption{A 2DoF Arm in an environment with obstacles that fully split the configuration space. 
    %making one of the valid goal solutions unreachable. 
    A naive seed for \eq{ik-opt} would produce Goal 2, which is impossible to reach from the initial configuration. Sampling the \textbf{optimal} goal configuration is necessary for sampling-based planners to converge quickly. 
    %Even though they can both be generated by \eq{ik-opt}. 
    The configuration space $\mathcal{C}_\text{free}$ is represented in green.}
    \label{fig:split_js_fig}
\end{figure}

We consider settings in which a robotic manipulator is planning motions in configuration space with a position controller. 
The robotic manipulator starts in an initial joint configuration $q_0 \in \mathcal{C}_\text{free} \subseteq \mathcal{Q}$ and intends to reach $q_N \in \mathcal{C}_\text{free}$. Here, $\mathcal{C}_\text{free}$ is the set of collision-free joint states and $\mathcal{Q}$ is the set of all robot joint states. 
The robot achieves this motion according to the continuous mapping $\mathbf{q}: \mathbb{R}^+ \to \mathcal{Q}$:
\begin{equation}
\begin{split}
    \mathbf{q}(0) &= q_0
    \\
    \mathbf{q}(1) &= q_N
    \\
    \forall t \in [0, 1], \quad&
    \mathbf{q}(t) \in \mathcal{C}_\text{free}
\end{split}
\label{eq:motion-planning}
\end{equation}
where $\mathbf{q}$ maps a normalized time parameter to a robot state $q$ in the collision-free space $\mathcal{C}_\text{free}$.
We represent the space of feasible motion plans as $\mathbf{Q}$.
We note that the goal position $g$ provided to the robot is defined in \textit{task} space. That is, $g = f(q_N)$ where $f: \mathcal{Q} \to \mathcal{X}$ is the robot's forward kinematics function and $g \in \mathcal{X}$ is the task space goal.

\p{Optimal Motion Planning}
Many possible $\mathbf{q}$ satisfy \eq{motion-planning}. An \textit{optimal} plan $\mathbf{q}^\star$ is found through \eq{optimal-motion-planning}:
\begin{equation}
\mathbf{q}^\star = 
    \mathop{\arg\min}\limits_{\mathbf{q} \in \mathbf{Q}}
    J(\mathbf{q})
    \label{eq:optimal-motion-planning}
\end{equation}
where $J$ is an objective function that should be minimized, such as path length or energy expenditure.
In this work, we focus on high-level position-based planning.
More specifically, we focus on motion plans of the following form:
\begin{equation}
    \begin{aligned}
        \forall i \in \{0, \dots, N\}, \quad & \mathbf{q}(t_i) = q_i \\
        \forall t \in [t_i, t_{i+1}], \quad & \mathbf{q}(t) = q_i + \frac{t - t_i}{t_{i+1} - t_i} (q_{i+1} - q_i) \\
        \text{s.t.} \quad & t_0 = 0, \quad t_N = 1
    \end{aligned}
    \label{eq:ps1}
\end{equation}
where the robot linearly interpolates its movement through a set of waypoints $\{q_0, q_1, \dots, q_{N - 1}, q_N\}$. We do not consider differential constraints in this work (such as those arising from robot dynamics).

\p{RRT-Style Planners}
Rapidly-Exploring Random Trees (RRT, also referred to as Rapid-Random Trees) are a sampling-based method for exploring configuration spaces according to a metric $\rho$ \cite{lavalle1998rapidly}. This metric is typically taken to be the \textit{path cost}, similar to $J$ in \eq{optimal-motion-planning}.

RRT is designed to be a swift, single-query (i.e., not reusable) data structure and sampling scheme for searching high-dimensional spaces subject to algebraic and differential constraints \cite{lavalle2000birrt}. 
RRT achieved this by iteratively extending the tree to a randomly sampled state in the environment along a vector of $\epsilon$ magnitude. This process is referred to as \texttt{EXTEND}.

The probability of RRT converging to an optimal plan is $0$. The introduction of asymptotically-optimal planners such as RRT$^\star$ \cite{karaman2011rrtstar} showed that swift, single-query planners could be redesigned with \texttt{REWIRE} to produce optimal solutions without substantial overhead. However, using these asymptotically-optimal planners in isolation for motion in high-dimensional \textit{configuration} spaces is still suboptimal. 

\p{Inverse Kinematics Solvers}
Recall that we are interested in providing the goal position in \textit{task} space, not configuration space.
In manipulators (and robotics more broadly), the joint state $q$ is mapped to the task space pose $x \in \mathcal{X}$ through the forward kinematics function $f$:
\begin{equation}
    x = f(q)
    \label{eq:fk}
\end{equation}
In general, $f$ is highly nonlinear and not (analytically) invertible. Quickly finding $q$ such that $x = f(q)$ is a widely researched topic known as \textit{Inverse Kinematics} (IK) \cite{singh2021review, colome2012redundant, beeson2015trac}. 
The leading approach to generate IK solutions is solve the equivalent constrained optimization problem \cite{beeson2015trac}:
\begin{equation}
\begin{split}
    \min\limits_{q} \quad &E(q) = \frac{1}{2} \|f(q) - x\|^2_W + \frac{1}{2} \lambda \| q - q^\prime \|^2
    \\
    \text{s.t.} \quad &q_{\text{min}} \le q \le q_{\text{max}}
\end{split}
\label{eq:ik-opt}
\end{equation}
where $W$ and $\lambda$ are weights, $x$ is the target task pose, and $q^\prime$ is a \textit{seed} joint configuration. Typically, \eq{ik-opt} is solved via SQP \cite{boggs1995sequential} using the gradient:
\begin{equation}
    \nabla E(q) = \nabla f(q)^T W \left(f(q) - x\right) + \lambda \left(q - q^\prime\right)
\end{equation}
Recall that $f$ is highly nonlinear. $E(q)$ can be subject to many local minima, so finding the correct seed $q^\prime$ is critical. Oftentimes, the current joint configuration is used as the seed. We will show in \ref{sec:method} that this choice of seed is suboptimal; performance improves when $q^\prime$ is carefully chosen. 

\begin{algorithm}[t]
    \caption{Many-RRT$^\star$}\label{algo:method}
    \begin{algorithmic}
        \Require $f, \mathcal{C}_\text{free}$ \Comment{FK, Conf. Space}
        \State $(x, q) \gets \text{sample\_from\_KDTree}(\mathcal{C}_\text{free})$
        \State $q^\prime \gets \text{parallel\_IK\_solutions}(f, x, q)$
        \State $q^\prime \gets \text{downsample}(q^\prime)$
        \State $N \gets \text{dim}(q^\prime)$
        \State $\mathcal{G} \gets \{G^0, \ldots, G^N\}$ \Comment{Goal Trees}
        \State $G^{N+1} \gets \varnothing$ \Comment{Start Tree}
        \For{$i \in \{0, \ldots, \text{max iterations}\}$}
            \For{$j \in \{0,\ldots,N\}$} \Comment{Executed Parallel Async.}
                \State $\mathcal{G} \gets$ single\_iteration\_RRT$^\star(G^j, \mathcal{C}_\text{free})$ 
            \EndFor{}
            \State $v_i \gets \text{sample\_vertex}(\mathcal{C}_\text{free})$ \Comment{\eq{sample-vi}}
            \State $G^{N+1} \gets \text{EXTEND}(G^{N+1}, v_i)$ \Comment{See \cite{karaman2011rrtstar}}
            \If{$v_i \in \mathcal{G}$}
                \State $G^{N+1} \gets \text{CONN\_TREE}(G^{N+1}, \mathcal{G})$ \Comment{Par. Async.}
            \EndIf
            \State $G^{N+1} \gets \text{REWIRE}(G^{N+1})$ \Comment{See \cite{karaman2011rrtstar}}
        \EndFor{}
    \end{algorithmic}
\end{algorithm}
% \vspace{-1.0em}

\begin{figure*}[!ht]
    \centering
    \includegraphics[width=1.0\linewidth]{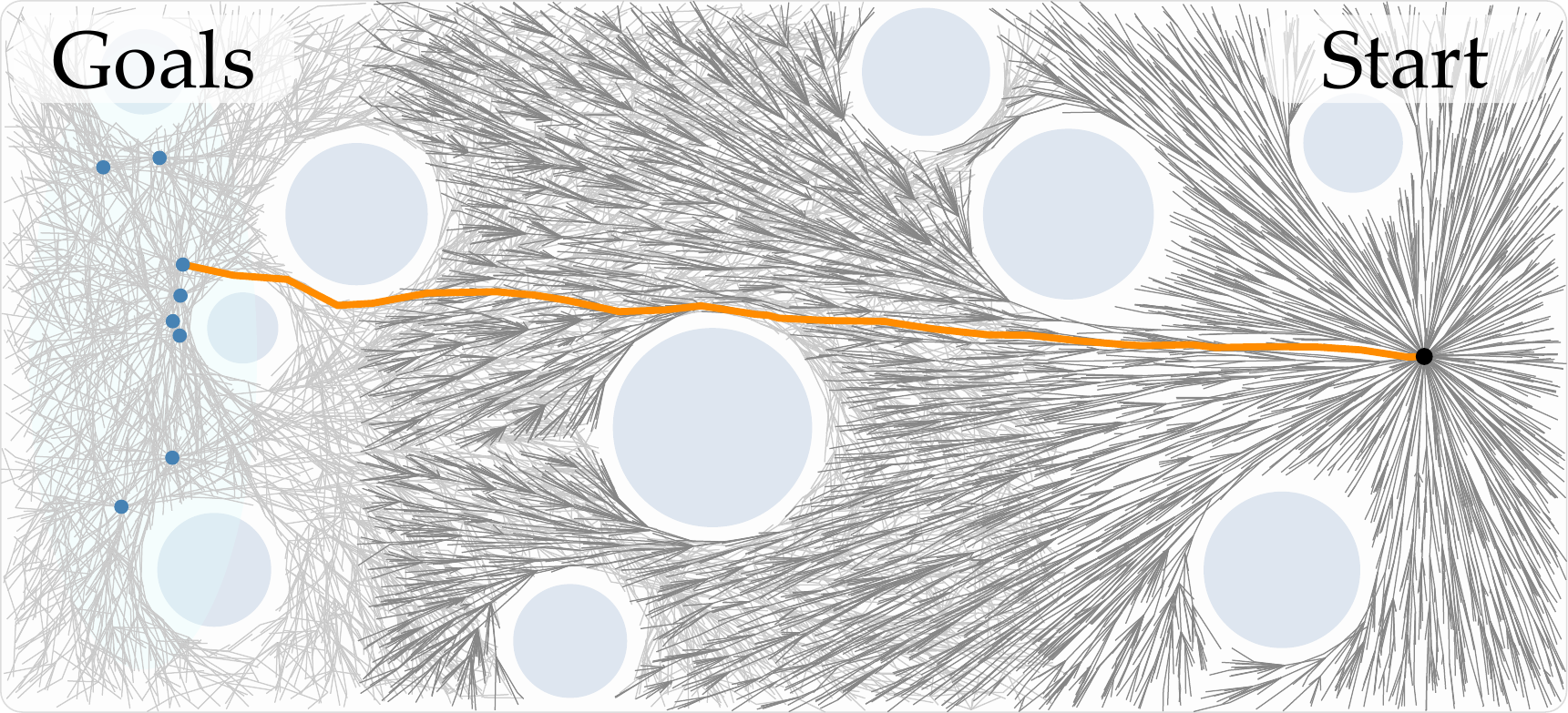}
    \caption{Many-RRT$^\star$ solves many bidirectional RRTs in parallel. 
    (a) We sample many points from the preimage of $g = f(q)$, which is a continuous manifold for redundant manipulators.
    (b) Asynchronously and in parallel, we execute a separate RRT$^\star$ thread for each configuration sampled in (a). These configurations are the \textit{roots} of these RRTs, and the start position is the goal.
    (c) We separately execute a RRT$^\star$ algorithm rooted at the start configuration that tries to connect to members of the trees in (b). This process is asynchronous, keeping track of the best solution yet given the state at each iteration.
    }
    \label{fig:method}
    \vspace{-1.5em}
\end{figure*}

Other implementations (such as the Orocos Kinematics and Dynamics Library \cite{kdl-url}) use the classic Newton-Rahpson iteration based on the pseudoinverse of the jacobian. Given $f$ and $q_0$, KDL iteratively performs the following:
\begin{equation}
\begin{gathered}
    J_q^k = \nabla f(q)\Big\rvert_{q = {q_k}}
    \\
    q_{k+1} = \text{proj}_{\mathcal{C}_\text{free}} \left(q_k + \alpha \left(J_q^k\right)^\dagger \left(x - f(q_k)\right)\right)
\end{gathered}
\label{eq:ik-opt2}
\end{equation}
until $\|q_{K + 1} - q_{K}\|^2 \le \epsilon$. This approach can be brittle near singularities, where the jacobian pseudoinverse becomes numerically unstable. Regardless, this IK solution is completely dependent on the initial configuration $q_0$, so it is subject to becoming trapped in local minima.

As previously mentioned, we are interested in finding optimal motion plans in \textit{configuration} space with $g \in \mathcal{X}$ as our goal pose. 
Without loss of generality, take $\mathcal{Q}_g = \{q \in \mathcal{Q} \mid g = f(q)\}$. 
Let $\mathbf{Q}_g = \{\forall q \in \mathcal{Q}_g, \exists \mathbf{q} \in \mathbf{Q} \mid \mathbf{q}(1) = q \wedge \mathbf{q} = \mathop{\arg\min}_{\mathbf{q}^\prime} J(\mathbf{q}^\prime) \}$ be the set of optimal motion plans for which the elements of $\mathcal{Q}_g$ are the final position. There is no guarantee that the set of plans $\mathbf{Q}_g$ have the same objective value. More precisely, $J(\mathbf{q})$ for each $\mathbf{q} \in \mathbf{Q}_g$ can be different. Thus, the optimal planning problem in configuration space reduces to a \textit{multi-arm bandit problem}, where the choice of $q \in \mathcal{Q}_g$ matters as much as the plan $\mathbf{q}$. Even if the plan $\mathbf{q}$ is optimal from $q_0$ to $q_N$, if the goal position $q_N$ is not ideal, then the original plan is suboptimal. 

Reasoning over the entire set $\mathcal{Q}_g$ is required to ensure global optimality.
However, while the forward kinematics function is a deterministic mapping, its inverse is generally ill-posed. For redundant manipulators, the preimage of a task-space configuration $g$ constitutes a continuous manifold of dimension $m - \dim(\mathcal{X})$, where $m$ is the number of degrees of freedom of the manipulator.\footnote{Even if $m \le \dim(\mathcal{X})$, there are often symmetries in the robot kinematics that can cause the preimage of a task-space configuration $g$ to have many possible joint-space configurations.}
The commonly used IK solutions presented in Equations~(\ref{eq:ik-opt}) and (\ref{eq:ik-opt2}) do not output a set of values: they produce singletons.
Consequently, an exhaustive search of the configuration space is computationally intractable due to the uncountable cardinality of the null space --- even with the IK solutions presented above.
Additionally, in the presence of obstacles, the configuration space of a robot can be fully split, meaning that a given singleton IK solution may be unreachable from the current configuration (as shown in \fig{split_js_fig}). 

%% file: 3_method.tex
\section{Many-RRT$^\star$}\label{sec:method}

Here we introduce our method for robust joint-space motion planning. Instead of finding a single feasible $q$ such that $g = f(x)$, with $g$ being the goal pose in task space, we rephrase the optimization problem from \eq{optimal-motion-planning} as:
\begin{equation}
    \begin{aligned}
        \mathbf{q}^\star &= \mathop{\arg\min}\limits_{\mathbf{q} \in \mathbf{Q}} J(\mathbf{q})
        \\
        \text{s.t.} \quad 
        \mathbf{q}(0) &= q_0
        \\
        \mathbf{q}(1) &\in \mathcal{Q}_g
        \\
        \forall t &\in [0, 1], \mathbf{q}(t) \in \mathcal{C}_\text{free}
    \end{aligned}
    \label{eq:m1}
\end{equation}
where instead of the decision variables being over the set $\{q_1, \ldots, q_{N - 1}\}$ with constraint $\mathbf{q}(1) = q_N$, we let $q_N$ be a decision variable as well, but subject $q_N$ to the constraint manifold $\mathcal{Q}_g$.\footnote{We note that the solution to \eq{m1} should also be of the same form as \eq{ps1}}
The solution to \eq{m1} is the \textit{globally optimal plan} in joint space. 

\p{Sampling IK Solutions}
As previously mentioned, solving for $\mathcal{Q}_g$ in realtime is infeasible.
Instead, we sample $K$ joint configurations from $\mathcal{Q}_g$ using the following two step process.
Prior to solving \eq{ik-opt}, we generate a large set of unlabeled joint configurations $M \subseteq \left(\mathcal{C}_\text{free}\right)^K$ via rejection sampling with a uniform distribution over $Q$ (in our experiments, we use $10^5$ samples).

This set $M$ can be computed once per robot morphology and reused across plans or interactions. 
We structure $M$ as a KD-Tree \cite{ram2019revisiting} for task- to joint-space querying using $K$-nearest neighbors. 
Once we have obtained $\{q^\prime_i\}_0^K$ seeds from $M$, we solve $K$ IK problems using the constrained optimization solver presented in \eq{ik-opt}:
\begin{equation}
\begin{split}
    \min\limits_{q} \quad &E(q) = \frac{1}{2} \|f(q) - x\|^2_W + \frac{1}{2} \lambda \| q - q^\prime_i \|^2
    \\
    \text{s.t.} \quad &q_{\text{min}} \le q \le q_{\text{max}}
\end{split}
\end{equation}
for each $i \in \{0, \ldots, K\}$. 
Empirically we find that using many possible seeds improves the convergence \eq{ik-opt} to minima it would not otherwise converge to. 
The resulting set of possible joint configurations is down-sampled to remove elements that satisfy $\|q_i - q_k\| \le \epsilon$ for each pair of IK solutions, where $\epsilon$ is parameter chosen by the designer (we use $1^{-4}$ in our experiments).
This down-sampled set $\{q^\prime_i\}_0^N$ serves as the set of possible \textit{goal} configurations for the $N$ parallel trees in what follows.

\begin{table*}[!th]
    \centering
    \small
    \renewcommand{\arraystretch}{0.85}
    \caption{Convergence Characteristics across $500$ trials for Many-RRT$^\star$ and baselines (Max Iteration: $3000$, Max Runtime: $3000$ms).}
    \label{tab:combined_stretched}
    \begin{tabular*}{\textwidth}{@{\extracolsep{\fill}} llcccccccc @{}}
        \toprule
        \textbf{Env.} & \textbf{Algorithm} & \textbf{Succ.} & \multicolumn{3}{c}{\textbf{Iteration of 1$^{\text{st}}$ Sol. ($\%$)}} & \multicolumn{2}{c}{\textbf{Median Cost}} & \multicolumn{2}{c}{\textbf{Avg. Runtime (ms)}}\\
        \cmidrule{4-6} \cmidrule{7-8} \cmidrule{9-10}
        && \textbf{Rate} & 10$^{\text{th}}$ & 50$^{\text{th}}$ & 90$^{\text{th}}$ & First & Final & First & Final \\
        \midrule
        \multicolumn{10}{c}{\textit{6-DoF Manipulator}} \\
        \midrule
        \multirow{3}{*}{Table} 
        & RRT$^\star$ & 98.2\% & \textbf{2.0} & \textbf{2.0} & \textbf{2.0} & \textbf{5.95} & 5.86 & \textbf{0.30} & 332.73 \\
        & RRT$^\star$-Connect  & 93.8\% & 73.0 & 147.5 & 597.4 & 9.96 & 7.41 & 31.87 & \textbf{261.82} \\
        & Many-RRT$^\star$ & \textbf{100.0\%} & 61.9 & 103.0 & 159.0 & 9.77 & \textbf{5.73} & 9.28 & 362.27 \\
        \addlinespace
        \multirow{3}{*}{Wall} 
        & RRT$^\star$ & \textbf{100.0\%} & 61.9 & 130.0 & 236.2 & 7.33 & 5.38 & 27.42 & 411.77 \\
        & RRT$^\star$-Connect & \textbf{100.0\%} & 412.6 & 751.0 & 1180.0 & 10.95 & 6.49 & 73.50 & \textbf{328.08} \\
        & Many-RRT$^\star$ & \textbf{100.0\%} & \textbf{25.0} & \textbf{42.0} & \textbf{101.0} & \textbf{4.81} & \textbf{4.06} & \textbf{5.42} & 472.49 \\
        \addlinespace
        \multirow{3}{*}{Passage} 
        & RRT$^\star$ & 93.6\% & \textbf{3.0} & \textbf{3.0} & \textbf{12.0} & \textbf{2.09} & 1.94 & 46.99 & 439.53 \\
        & RRT$^\star$-Connect & \textbf{100.0\%} & 13.9 & 28.0 & 68.0 & 2.35 & \textbf{1.81} & \textbf{2.69} & \textbf{336.95} \\
        & Many-RRT$^\star$ & \textbf{100.0\%} & 11.0 & 24.0 & 73.0 & 2.41 & \textbf{1.81} & 3.41 & 918.49 \\
        \addlinespace
        \multirow{3}{*}{Random} 
        & RRT$^\star$ & 1.4\% & >3000 & >3000 & >3000 & $\infty$ & $\infty$ & 448.07 & 454.48 \\
        & RRT$^\star$-Connect & 1.6\% & >3000 & >3000 & >3000 & $\infty$ & $\infty$ & 324.91 & \textbf{322.86} \\
        & Many-RRT$^\star$ & \textbf{100.0\%} & \textbf{135.9} & \textbf{225.0} & \textbf{468.3} & \textbf{16.57} & \textbf{12.97} & \textbf{41.98} & 569.91 \\
        \midrule
        \multicolumn{10}{c}{\textit{7-DoF Manipulator}} \\
        \midrule
        \multirow{3}{*}{Table} 
        & RRT$^\star$ & \textbf{100.0\%} & \textbf{2.0} & \textbf{2.0} & \textbf{7.0} & 5.92 & 5.89 & \textbf{0.71} & 517.46 \\
        & RRT$^\star$-Connect  & \textbf{100.0\%} & 43.0 & 60.0 & 90.0 & 7.62 & 6.29 & 7.41 & \textbf{382.89} \\
        & Many-RRT$^\star$ & \textbf{100.0\%} & 24.0 & 34.0 & 49.1 & \textbf{5.17} & \textbf{3.31} & 4.30 & 434.98 \\
        \addlinespace
        \multirow{3}{*}{Wall} 
        & RRT$^\star$ & \textbf{100.0\%} & \textbf{8.0} & 129.0 & 348.4 & 9.38 & 8.32 & 52.89 & 748.32 \\
        & RRT$^\star$-Connect & \textbf{100.0\%} & 58.8 & 229.0 & 593.4 & 9.66 & 8.34 & 43.34 & \textbf{550.24} \\
        & Many-RRT$^\star$ & \textbf{100.0\%} & {39.0} & \textbf{54.0} & \textbf{91.0} & \textbf{7.16} & \textbf{5.27} & \textbf{10.36} & 764.34 \\
        \addlinespace
        \multirow{3}{*}{Passage} 
        & RRT$^\star$ & 58.1\% & \textbf{9.0} & 368.5 & >3000 & 8.82 & 7.40 & 376.07 & 744.12 \\
        & RRT$^\star$-Connect & 57.2\% & 72.0 & 337.0 & >3000 & 9.33 & 7.02 & 254.73 & \textbf{534.55} \\
        & Many-RRT$^\star$ & \textbf{100.0\%} & 48.0 & \textbf{78.0} & \textbf{186.2} & \textbf{7.50} & \textbf{5.81} & \textbf{15.27} & 675.10\\
        \addlinespace
        \multirow{3}{*}{Random} 
        & RRT$^\star$ & 5.0\% & >3000 & >3000 & >3000 & $\infty$ & $\infty$ & 1799.51 & \textbf{1747.87} \\
        & RRT$^\star$-Connect & 8.8\% & >3000 & >3000 & >3000 & $\infty$ & $\infty$ & 2880.15 & 2843.35 \\
        & Many-RRT$^\star$ & \textbf{96.8\%} & \textbf{430.6} & \textbf{904.0} & \textbf{2878.0} & \textbf{14.36} & \textbf{7.25} & \textbf{1154.90} & 2479.62 \\
        \bottomrule
    \end{tabular*}
    \vspace{-1.0em}
\end{table*}

\p{Parallelizing Tree Generation}
Core to our method is a set of directed graphs (trees) $\{G^0, G^1, \ldots, G^N, G^{N+1}\}$ with $G^i = \{V^i, E^i\}$ where $V^i$ is the set of vertices and $E^i$ is the set of edges. 
The set $\{G^0, \ldots, G^N\}$ are trees that grow from each \textit{goal} configuration to the \textit{start} configuration using standard RRT$^\star$ \cite{karaman2011rrtstar}.\footnote{In practice, we use a much lower sampling threshold for these trees. In traditional RRT$^\star$, a sampling ratio of $0.5$ is used. We use a sampling ratio much closer to $0$ to bias \textit{exploration} instead of \textit{exploitation}.}
These trees grow in parallel on separate threads, completely independently. 
The tree $G^{N+1}$ is rooted at the \textit{start} configuration. At each iteration $i$, $G^{N+1}$ extends to a node according to the following sampling rule:
\begin{equation}
v_i \sim
\begin{cases}
    U\left[\mathcal{C}_\text{free}\right] & \gamma > \gamma_0
    \\
    U\left[\{q^{\prime}_k\}_0^N \cup \bigcup_{k=0}^{k=N}\{E_i^k - E_{0:i - 1}^k\}\right] & \text{else}
\end{cases}
\label{eq:sample-vi}
\end{equation}
where $\gamma \sim U[0, 1]$ and $\gamma_0 \in (0, 1)$. Put another way, the tree $G^{N+1}$ explores the joint space with probability $\gamma$. With probability $1 - \gamma$, the tree extends to new nodes added to the other trees or any of their goal configurations. This sampler ensures that $G^{N+1}$ balances exploration with exploitation.

\p{Implementation Details}
Our method is summarized in Algorithm~\ref{algo:method} and \fig{method}. 
Multithreading is advantageous to our method. Each of the trees rooted at the goal positions are independent, so their growth is trivially parallelizable. 

We find that the worst case running time of our current implementation is $O(mn^2)$ (where $m = N + 1$ and $n = \text{nodes}_\text{max}$).
While the extension of the tree is $O(mn\log{n})$, we find that through parallelization, the time complexity reduces to $O(n\log{n})$ (assuming $n_\text{proc} \geq N + 2$), matching time complexity of RRT$^\star$.
However, the bound is dictated by the costly \texttt{CONN\_TREE} method, which must compare $n$ nodes in the start tree to $n$ nodes in each of the $m$ goal trees. 
% Omit below? %
Future iterations may aim to further parallelize the \texttt{CONN\_TREE} method, but the parallel slackness of the method is very high, meaning that many available processors would be needed to achieve the speedup.

In our testing, we implemented multithreading on a CPU, but this approach could be trivially extended to use GPU or NPU.

%% file: 4_study.tex
\begin{figure}
    \centering
    \includegraphics[width=1.0\linewidth]{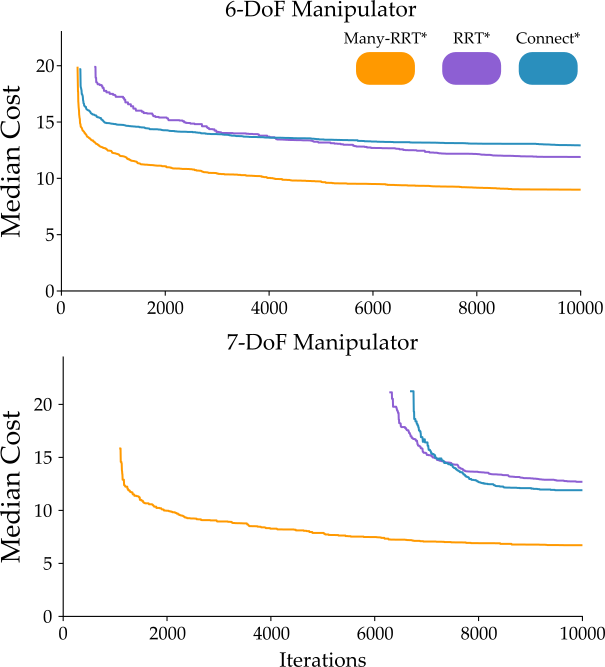}
    \caption{The median cost per iteration across trials for the \textit{Random} environment with $500$ trials per method. Not only does Many-RRT$^\star$ converge to a feasible path before baselines, but it also finds lower-cost trajectories as the number of iterations increases.
    The performance difference widens as the dimension of the configuration-space increases: for the 7-DoF manipulator, Many-RRT$^\star$ finds a feasible solution in $6$ times fewer iterations.
    }
    \label{fig:cpi}
    \vspace{-1.0em}
\end{figure}
\begin{figure}
    \centering
    \includegraphics[width=1.0\linewidth]{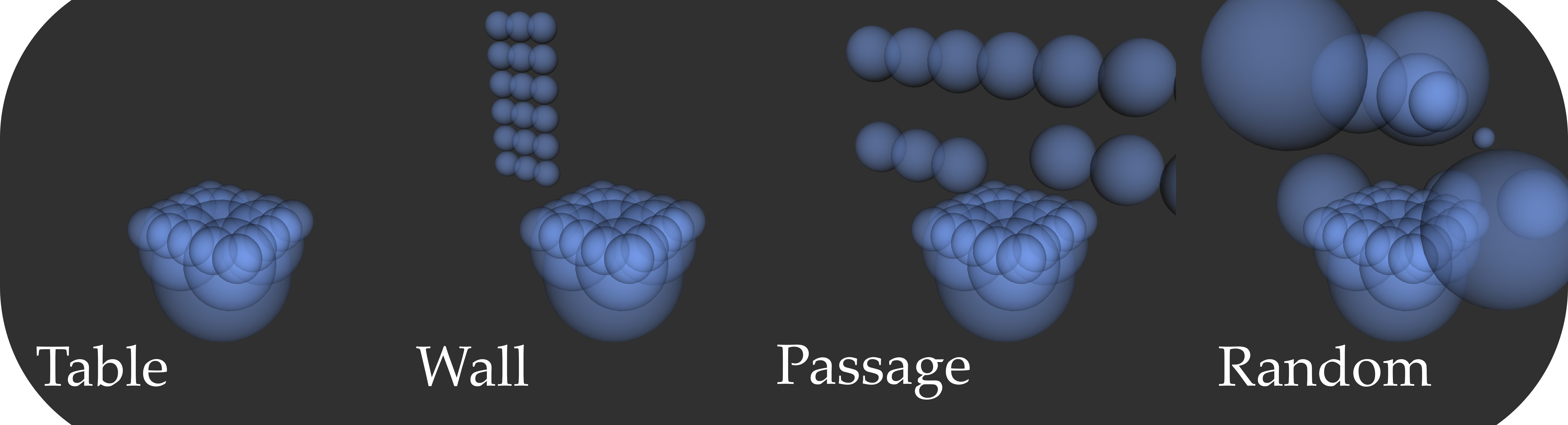}
    \caption{The collision representations of the environments used in our simulated experiments.  
    From left to right, the environments increase in relative difficulty.
    In \textit{Table}, the manipulator must move from one pose to another while avoiding the flat table in the center of the frame.
    In \textit{Wall}, the manipulator must move from one side to another of the wall while avoiding both the table and the wall. 
    Likewise, in \textit{Passage}, the manipulator must move from one side to the other by navigating through the passage. In this environment, the strengths of bi-directional sampling-based planners becomes more evident, as methods like RRT$^\star$-Connect are able to explore both areas to attempt to connect the joint-space through the narrow passage.
    In the most challenging environment, \textit{Random}, a random configuration of spherical objects are scattered throughout the environment. 
    Across environments, we ensure that both the goal and start positions are collision-free and that a feasible motion plan exists before experimentation.
    }
    \vspace{-1.5em}
    \label{fig:environments}
\end{figure}
\section{Experiments}\label{sec:study}

To compare the algorithms, we tested the performance of each algorithm in creating path plans through a variety of environments on 6DoF and 7DoF arms (UR10e and Franka Panda, respectively).

All of the environments used are represented as spheres for compatibility with our collision algorithm, which based off of the approach seen in \cite{sundaralingam2023curobo}. We developed four test environments, shown in \fig{environments}. 
The environments are scaled to the advertised reach of the robot at full extension $l$ (0.85m for the Franka Panda and 1.3m for the UR10e), with their origin being the base link of the arm. 

Each environment was designed to compare the performance of the algorithms in a different way.
The table is meant to show performance in largely free space as a baseline.
The wall is meant to show performance with a simple obstacle, meaning that some IK solutions will have a straight-line path from start to goal in the joint space, while other solutions will require more complex planning. 
The passage is meant to show performance with a fully bifurcated configuration space, that is, there are some valid IK solutions at each goal that are simply unreachable from certain IK solutions at the start pose.
Finally, the random environment is meant to show a very difficult environment to plan through in order to stress-test the algorithms.

We compared the performance of RRT$^\star$, RRT$^\star$-Connect, and Many-RRT$^\star$ on 6DoF and 7DoF arms (UR10e and Franka Panda).
For the multi-threaded algorithms --- RRT$^\star$-Connect and ManyRRT$^\star$ --- time benchmarking is measured from the beginning of the first thread starting to the end of the tree ascension (when a path is created). It does not include data structure initialization or destruction. 
For the single threaded RRT$^\star$, time benchmarking is --- similarly --- measured from the beginning of the tree extension loop until the end of the tree ascension.

Benchmarking is done over 500 trials, with task-space start and goal poses for each trial being the same for all three algorithms.
Benchmarks for \textit{Wall} and \textit{Passage} trials used the same goal poses for both arms\footnote{Scaled to the advertised length of the arm}, whereas those for \textit{Random} and \textit{Table} trials used poses based on each arm's reachable positions in the \textit{Random} environment.

All convergence characteristic trials were performed using $\text{nodes}_\text{max} =$ 3000 with 3000ms timeout. 
We completed all trials for Many-RRT$^\star$ using 10 IK solutions, resulting in 11 trees being built.
Two sets of convergence characteristic trials were performed for each platform, with one exiting on first success and the other running until max nodes was reached. The results of the convergence rate trials for the 6DoF and 7DoF arms are seen in Table~\ref{tab:combined_stretched}.

Importantly, the maximum nodes for Many-RRT$^\star$ is dictated by the sum of any tree with the start tree. This means that any call to Many-RRT$^\star$ can have up to $\frac{\text{nodes}_\text{max}}{2} (N+1)$ nodes.
As a consequence of this, we define an \textit{iteration} $i$ at any given time $t$ in terms of the number of nodes and trees: $i_t = \lfloor \sum_{k=0}^{N+1} 2\cdot\text{nodes}(k)_t / (N + 1) \rfloor$. This ensures that Many-RRT$^\star$ will conform to the given \textit{iterations}, though it is allowed more \textit{nodes}.

Benchmarks were performed on an AMD Ryzen 9 7950X CPU with virtual multithreading enabled and $128$GB of RAM operating at a clock rate of $4800$MHz.

In addition to convergence rate testing, we performed comparative tests to characterize the best cost at each iteration for each algorithm (cost-per-iteration). These trials were performed in the \textit{Random} environment to highlight the discrepancy between each algorithm. The results of these trials are seen in \fig{cpi}. These graphs show the median cost per iteration over $500$ trials, each using $10^4$ iterations.

\p{Results}
In the convergence rate trials, we find that while RRT$^\star$ and RRT$^\star$-Connect perform on-par with Many-RRT$^\star$ in simple environments such as \textit{Table} and \textit{Wall}, Many-RRT$^\star$ significantly outperforms in terms of median cost in complex environments such as \textit{Passage} and \textit{Random}.

In the cost-per-iteration trials, we find not only that Many-RRT$^\star$ consistently converges sooner than RRT$^\star$-Connect or RRT$^\star$, but also that it converges to a significantly lower cost than either baseline. 

%% file: 5_conclusion.tex
\section{Conclusion}\label{sec:conc}

In this paper we discuss challenges inherent when creating optimal trajectories for high DoF serial manipulators. We address these challenges and offer an asymptotically globally-optimal algorithm for planning in the configuration space of serial manipulators. 
Our key contribution is parallelizing bi-directional exploration to \textit{many} valid inverse kinematics solutions to address the non-invertability of forward kinematics when planning in the configuration space.

In practice, we find that the proposed approach offers more robust \textit{and} optimal planning than previous iterations of the RRT algorithm when planning in the configuration space of high-DoF serial manipulators, and does so without significantly compromising on runtime. 